\documentclass[conference]{IEEEtran}
\IEEEoverridecommandlockouts
\usepackage{cite}
\usepackage{amsmath,amssymb,amsfonts}
\usepackage{algorithmic}
\usepackage{graphicx}
\usepackage{textcomp}
\usepackage{xcolor}
\usepackage[hyphens]{url}

\def\BibTeX{{\rm B\kern-.05em{\sc i\kern-.025em b}\kern-.08em
    T\kern-.1667em\lower.7ex\hbox{E}\kern-.125emX}}
\begin{document}

\title{Position Paper: \\ Metadata Enrichment Model: Integrating Neural Networks and Semantic Knowledge Graphs for Cultural Heritage Applications}

\author{
\IEEEauthorblockN{Jan Ignatowicz}
\IEEEauthorblockA{\textit{Faculty of Physics, Astronomy} \\ \textit{and Applied Computer Science} \\
\textit{Jagiellonian University}\\
Kraków, Poland \\
0009-0005-8716-4175}
\and
\IEEEauthorblockN{Krzysztof Kutt}
\IEEEauthorblockA{\textit{Faculty of Physics, Astronomy} \\ \textit{and Applied Computer Science} \\
\textit{Jagiellonian University}\\
Kraków, Poland \\
0000-0001-5453-9763}
\and
\IEEEauthorblockN{Grzegorz J. Nalepa}
\IEEEauthorblockA{\textit{Faculty of Physics, Astronomy} \\ \textit{and Applied Computer Science} \\
\textit{Jagiellonian University}\\
Kraków, Poland \\
0000-0002-8182-4225}
}

\maketitle

\begin{abstract}
The digitization of cultural heritage collections has opened new directions for research, yet the lack of enriched metadata poses a substantial challenge to accessibility, interoperability, and cross-institutional collaboration. In several past years neural networks models such as YOLOv11 and Detectron2 have revolutionized visual data analysis, but their application to domain-specific cultural artifacts - such as manuscripts and incunabula - remains limited by the absence of methodologies that address structural feature extraction and semantic interoperability. In this position paper, we argue, that the integration of neural networks with semantic technologies represents a paradigm shift in cultural heritage digitization processes. We present the Metadata Enrichment Model (MEM), a conceptual framework designed to enrich metadata for digitized collections by combining fine-tuned computer vision models, large language models (LLMs) and structured knowledge graphs. The Multilayer Vision Mechanism (MVM) appears as the key innovation of MEM. This iterative process improves visual analysis by dynamically detecting nested features, such as text within seals or images within stamps. To expose MEM’s potential, we apply it to a dataset of digitized incunabula from the Jagiellonian Digital Library and release a manually annotated dataset of 105 manuscript pages. We examine the practical challenges of MEM's usage in real-world GLAM institutions, including the need for domain-specific fine-tuning, the adjustment of enriched metadata with Linked Data standards and computational costs. We present MEM as a flexible and extensible methodology. This paper contributes to the discussion on how artificial intelligence and semantic web technologies can advance cultural heritage research, and also use these technologies in practice.
\end{abstract}


\section{Introduction}
The cultural heritage collections' digitization has transformed the way, how global audiences may access historical artifacts and explore manuscripts, artworks, and archaeological findings. The GLAM \cite{vanhooland_facet_2014} institutions, which are galleries, libraries, museums, and archives now makes available increasingly vast digital repositories \cite{10.1145/3607542.3617354}. They spread knowledge and foster cross-disciplinary research. However, the true potential of such collections remains limited by the lack of enriched and semantically structured metadata. Traditional metadata schemas, such as Dublin Core \cite{Weibel1998} or MARC \cite{guenther2003marc}, stand out at presenting basic data (e.g. title, author, date). However, they often fail in representing the complexity of cultural artifacts \cite{Doerr2003}. For instance, medieval manuscripts may consists of complex seals, handwritten marginalia, or ornamentations that contain historical context, which are the elements rarely documented in existing metadata \cite{vanhooland_facet_2014}. This gap not only limits scholarly analysis but also hinders interoperability and integration with Linked Data ecosystems \cite{berners2006linked}, which are getting more and more valuable for modern digital humanities.

Recent advances in artificial intelligence and neural networks \cite{10.1007/s00371-024-03343-0} offer interesting tools that can be used for preparing automated metadata enrichment mechanisms. Models like YOLO \cite{Khanam2024} , Detectron2 \cite{wu2019detectron2} or HRNet \cite{wang2020deep} have determined notable success in object detection and segmentation. On the other hand large language models (LLMs) stand out at parsing and generating text descriptions \cite{bommasani2021opportunities}. Nevertheless, to the best of our knowledge, they are not yet widely used in cultural heritage applications \cite{Garcia-Garcia2017}. Most approaches focus on isolated tasks, such as detecting text regions or classifying visual themes, without addressing the structural nature of historical artifacts \cite{shi2023gold}. For example, a 15th-century incunabulum, may consist of a seal containing text, which may indicate its origin. The nested structure requires iterative analysis. Furthermore, so far, few solutions integrate AI-driven insights with semantic knowledge graphs \cite{liu2024knowledge}, which is crucial for creating interoperable and Linked Data-wise metadata. Existing frameworks like Europeana’s Data Model (EDM) \cite{isaac2009europeana} or CIDOC CRM \cite{Doerr2003} provide a reliable ontological basis but lack mechanisms to dynamically incorporate AI-generated, or rather AI-retrieved, knowledge.

In this position paper, we propose the Metadata Enrichment Model (MEM), a holistic framework designed to deal with discussed issues. MEM combines three core components: (1) retrained computer vision models for detecting or segmenting visual elements \cite{chen2017rethinking}, (2) large language models for contextual analysis and decision-making \cite{openai2023gpt4}, and (3) semantic knowledge graphs for structuring enriched metadata \cite{hong2024semantic} along with provided existing metadata. The key module of MEM is the Multilayer Vision Mechanism (MVM). This iterative visual analysis workflow dynamically uncover nested features. For example, MVM might first detect a seal on a manuscript page, then extract text within that seal \cite{smith2023ocr} in order to finally link the text to external knowledge bases like Wikidata \cite{vrandecic2014wikidata}. This mechanism is guided by the usage of LLMs in decision-making process. The MEM approach not only enrich metadata but also link them to external databases by creating knowledge graphs \cite{wilkinson2016fair}, enabling cross-institutional querying and integration.

To validate MEM’s practicability, we conduct a proof-of-concept study on a dataset of digitized incunabula from the Jagiellonian Digital Library \cite{jdl}. As part of this work, we release a manually annotated dataset of 105 manuscript pages. We marked visual elements such as seals, paragraphs, and decorative initials in ten distinct categories. We discuss the challenges of retraining neural networks models on domain-specific data \cite{tan2019efficientnet}. We indicate the role of semantic graphs in resolving ambiguities (e.g. distinguishing similar seals from different historical periods) \cite{vanhooland_llc_2015} and the implications of applying MEM in GLAM institutions. By examining technical barriers - such as computational costs, model generalization \cite{he2016deep}, and ontology adjustment this paper contributes to the growing discussion on the implementation and limitations of AI in cultural heritage digitization.

Our primary contributions are threefold:
\begin{itemize}
  \item The Metadata Enrichment Model (MEM), a conceptual framework that integrates neural networks, semantic technologies, and iterative analysis for metadata enrichment.
  
  \item The Multilayer Vision Mechanism (MVM), a dynamic workflow for finding nested features in cultural artifacts.
  
  \item An open dataset of manually annotated manuscripts to support future research in cultural heritage applications.
\end{itemize}

The following paper is structured as follows: Section 2 reviews related work in AI-driven metadata enrichment and semantic technologies. Section 3 presents the MEM framework along with MVM. Section 4 shows the case study, dataset, and proof-of-concept results. Section 5 discusses challenges and future directions. Finally section 6 concludes the paper.

\section{Related Work}

\subsection{Metadata Enhancement Progress}
The growing need for richer metadata in cultural heritage collections has evolved from early standardizations to nowadays approaches driven by AI. Initiatives like Dublin Core \cite{Weibel1998} and CIDOC CRM \cite{Doerr2003} shown basis schemas of metadata organization. Linked Data principles \cite{BernersLee2006} developed interoperability standards. Projects such as Europeana \cite{Purday2009} proved the potential of semantic technologies in connecting with global repositories. However, their trust in manual management is limited by scalability. Recent work by \cite{knowledge-graph-visualization} indicate the growing need for automating metadata workflows, but notes the absence of solutions for nested visual elements in artifacts like manuscripts - a gap MEM directly addresses.

\subsection{Computer Vision in Cultural Heritage}
Pre-trained models like YOLOv11 \cite{Khanam2024} and Detectron2 \cite{wu2019detectron2} were created for object detection in common contexts. Their application to historical documents may be testes but this activity requires some domain-specific adjustments. Models designed for semantic segmentation such as DeepLabv3 \cite{chen2017rethinking} and HRNet \cite{wang2020deep} might be used to find regions of interest, such as marginalia or seals, in digitized manuscripts. However, Garcia-Garcia \cite{Garcia-Garcia2017} noted, that these methods often struggle with structured data like text embedded in seals or annotations in illustrations. MEM’s Multilayer Vision Mechanism enters here by iteratively conducting multi-layer visual analysis with AI-guided help.

\subsection{LLMs and Hybrid Workflows}
The rise of large language models (LLMs) like GPT-4 \cite{openai2023gpt4}, oLLama \cite{ollama} and DeepSeek v3 \cite{deepseekv3} has opened new directions for metadata generation. Studies such as "On the Opportunities and Risks of Foundation Models" \cite{bommasani2021opportunities} accent their potential in multimodal applications and articles like: \cite{10.1007/978-3-031-77844-5_9}, \cite{chepurova-etal-2023-better}, \cite{Lippolis2023EnhancingEA}, \cite{d3159688d57f4176808086ec1de5377a}, \cite{lairgi:hal-04607294} and \cite{SHIMIZU2025100862} show how synergies between LLMs and semantic wikis like Wikidata can be made. However, as presented in articles \cite{10.1145/3581783.3611753} and \cite{10.1609/aaai.v37i9.26357} it is shown, that most frameworks treat text and visual analysis as separate tasks. MEM comes forward this division by integrating LLMs into its workflow as decision-makers for iterative vision tasks and as tools for contextualizing extracted metadata into knowledge graphs.

\subsection{Linked Data and Challenges}
Although some progress has been made in Linked Data implementations (e.g. the British Library’s \cite{britishlibrary2011bnb} linked open data project), practical difficulties still exist. In book \cite{vanhooland_facet_2014} authors identify inconsistencies in metadata quality as key obstacles, particularly for cross-institutional research. Recent work \cite{liu-etal-2024-knowledge-graph} proposes ontology growth using LLMs, but omit the role of visual data. MEM propose automated creation of ontology based on multimodal inputs like text and images. Finally, linking results to external repositories like Wikidata.

\subsection{MEM’s Contributions to Nowadays Gaps}
Existing approaches often spotlight isolated tasks like only detecting main objects, generating text or structuring metadata. It is done without connecting these processes each other. For instance, authors of \cite{li2024handwritingrecognitionhistoricaldocuments} improve structural segmentation in historical documents, but they neglect the issue of semantic interoperability. Similarly, Europeana’s EDM standardizes metadata but lacks mechanisms for enrichment supported by AI methods. MEM addresses these limitations through:

\begin{figure*}[t]
    \centering
    \includegraphics[width=\textwidth]{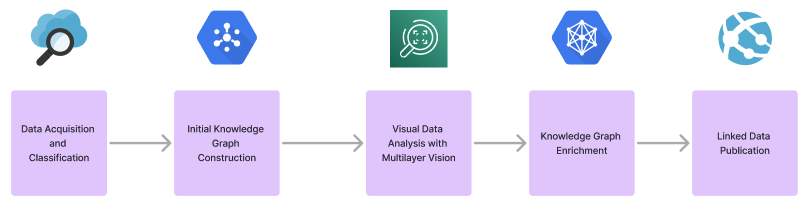}
    \caption{Metadata Enrichment Model (MEM) main flow.}
    \label{fig:mem_main_flow}
\end{figure*}

\begin{itemize}
    \item Multilayer Vision Mechanism (MVM): Iterative detection of nested features (e.g. text within seals) guided by LLMs.

    \item Hybrid Integration: Combining retrained CV models, LLMs, and semantic knowledge graphs.

    \item Dynamic Ontology Expansion: Auto-updating ontologies in RDF format based on extracted data.
\end{itemize}

By combining these innovations, MEM offers a scalable, interoperable framework for metadata enrichment. This stands as essential step toward intelligent digital heritage systems.


\section{The Metadata Enrichment Model (MEM)}
The Metadata Enrichment Model (MEM) is a modular conceptual framework designed to address the shortcomings of retrieving metadata in cultural heritage collections in traditional ways. MEM integrates neural networks, semantic technologies, and iterative analysis to dynamically enrich metadata. It also ensures interoperability with Linked Data ecosystems. The model’s architecture is interchangeable. It allows entering various modalities, like images, text, audio, and geospatial data. However, the presented here implementation focuses on incunabula and manuscripts as a proof of concept. The main MEM flow is presented in Figure~\ref{fig:mem_main_flow}.

A key challenge in metadata enrichment is the representation of complex visual and text elements in a semantic graph format. Traditional metadata schemas consist of basic bibliographic attributes and often fail in describing complex details such as marginalia, seals, handwritten text, or artistic decorations. MEM addresses this gap by usage of computer vision models in order to detect and classify these features. Then, structured knowledge graph encodes the extracted information and integrates them into a broader linked open data ecosystem.

\subsection{Multimodal Input Support}
\begin{figure}[!ht]
    \centering
    \includegraphics[width=0.45\textwidth]{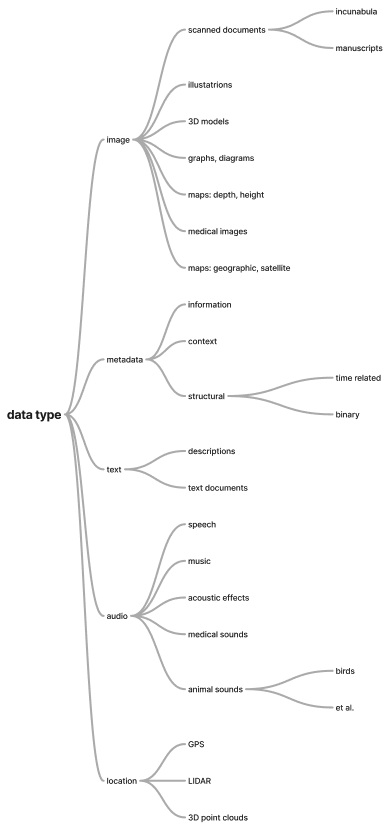}
    \caption{Possible entry modalities for MEM.}
    \label{fig:modalities}
\end{figure}

MEM’s architecture contain various modalities as presented in Figure~\ref{fig:modalities}. But the current use case indicate visual analysis of incunabula. Text metadata, like author and date, are processed by language models to create structured descriptions. in future extensions, audio recordings like oral histories could be analyzed via speech-to-text models. Geospatial data, such as the origin of artifacts, could be encoded using the W3C Geo vocabulary. This flexibility ensures that MEM can evolve due to changeable digitization trends.

\subsection{Semantic Knowledge Graph Construction and Dynamic Expansion}
MEM structures enriched metadata with RDF (Resource Description Framework), accordingly to Linked Data principles. Instead of relying on rigid ontologies like CIDOC CRM, the framework dynamically enhance schemas customizing them to the dataset’s characteristics. For example, if MVM detects a previously unclassified type of seal, MEM extends the ontology by defining a new class (e.g. \textit{mem:DecorativeSeal}) and linking it to existing entities via properties like \textit{mem:containsText} or \textit{mem:locatedInMargin}.

Actually conducted RDF - based ontology within project Cultural Heritage Exploration and Retrieval with Intelligent Systems (CHExRISH) at the Jagiellonian University, which focuses on historical artifacts - may also serve as a basic template in domain-specific use case. MEM can handle with such efforts by publishing enriched metadata via SPARQL endpoints, enabling combined queries across institutions. For instance, a seal detected in a Jagiellonian Library manuscript, might be linked to a Wikidata entry with its historical owner. That would create a bridge between local and global knowledge graphs.

\subsection{Multilayer Vision Mechanism (MVM)}
\begin{figure}[!ht]
    \centering
    \includegraphics[width=0.3\textwidth]{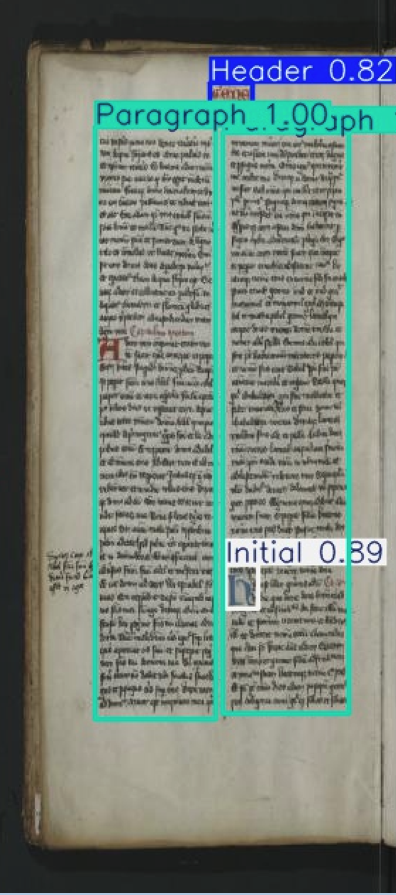}
    \caption{An example image from the created dataset. The image shows the detected classes: \textit{initial}, \textit{header}, and \textit{paragraph}.}
    \label{fig:detected_object}
\end{figure}

The Multilayer Vision Mechanism (MVM) is the key module of MEM's framework in image understanding processes. It is designed to handle with the structured nature of cultural relics. MVM iteratively processes visual data in order to uncover nested features. For example, a digitized manuscript page might first pass object detection, using YOLO or Detectron2, to identify high-level elements like seals or paragraphs. An example image from dataset with objects detected by the retrained YOLOv11 model is presented in Figure~\ref{fig:detected_object}. Regions of interest are then cropped and analyzed by further specialized models - such as HRNet for segmentation or Tesseract OCR for text extraction and understanding. That should reveal minor details, like text within a seal or annotations in margins.

The decision to continue or stop iteration is conducted by a hybrid approach:

\begin{itemize}

    \item Image-to-Text Conversion: A language model like BLIP-2 generates a text description of the cropped region (e.g. “A circular seal with Latin text along the border”).

    \item LLM-Conducted Decision: A language model like Mistral-7B evaluates the description with predefined queries, such as “Does this region contain text requiring transcription?” The LLM’s output (YES or NO) is combined with rule-based checks, like minimum pixel size or confidence thresholds. That determines whether further analysis is needed.
    
\end{itemize}

This process repeats until no new semantically meaningful features are detected or a maximum iteration depth (e.g. 3 layers) is reached. By separating detection or segmentation models from decision logic, MEM remains adaptable. It can swap YOLO for Mask R-CNN and at the same time requires minimal changes to the workflow.

\subsection{Integration with Linked Data}
The final phase of MEM results in makins enriched metadata accessible, interoperable, and linked with external knowledge sources. The knowledge graphs created in the previous stage are published as Linked Open Data (LOD) resource. It enables seamless integration with existing databases and cultural heritage repositories.

To achieve this, MEM provides retrieved metadata via SPARQL endpoints. That enables researchers to execute advanced queries and retrieve structured information about manuscripts. Additionally, MEM links detected entities, such as authors, institutions, or historical locations, with external resources like Wikidata and DBpedia. Through that it provides contextual depth and allows integrations across institutions.

Ensuring compatibility with FAIR \cite{wilkinson2016fair} principles (Findable, Accessible, Interoperable, Reusable) is an essential aspect of MEM’s architecture. Structuring metadata in a way according to semantic web standards, MEM facilitates interoperability between different institutions and collections. It allows the researchers to conduct comparative analyses and find interconnections between documents from various sources.

\subsection{The Role of Large Language Models}
LLMs in MEM serve a dual purpose: (1) Contextual Decision-Making. It analyzes text descriptions of visual regions and determine whether deeper iteration is justified. For instance, a seal described as “faded with illegible text” might trigger a stop signal, whereas “text in Gothic script” evoke some OCR model to use \cite{gao2023enabling}, \cite{zhang2024longcite}; (2) Metadata Generation. LLMs combine information from multiple layers (e.g. “The seal refers to King Charles V, as indicated by the transcribed text ‘CAROLUS REX’”) and map it into created ontologies \cite{liu2024knowledge}.

To mitigate hallucinations, MEM may employ a hybrid approach. Firstly using rule-based guards. Hard limits on iteration depth, region size, and confidence scores \cite{kazemi2025refining}, \cite{li2024rulerimprovingllmcontrollability}. Secondly, fine-tuning smaller LLMs (e.g. Mistral-7B) on domain-specific Q\&A pairs to improve reliability \cite{zhang2024longcite}, \cite{li2024rulerimprovingllmcontrollability}.

\subsection{Implications and Scalability}

The MEM framework is designed for being scalable. With more and more collections being processed, MEM can be adjusted to new use cases. Its modular architecture allows for future extensions, such as integration 3D object recognition for historical artifacts or expantion to audio processing for spoken heritage recordings.

Furthermore, MEM’s methodology can be used in other than historical manuscripts. Museums, archives, and academic institutions can use MEM for automated metadata retrieval, what may start new applications in digital conservation, origin tracking, and data-driven historical analysis.

By presenting MEM as an adjustable and scalable metadata enrichment methodology, we combine traditional cataloging approaches with AI-driven analysis. It makes cultural heritage collections more discoverable, structured, and interconnected. In the next section MEM is presented in action in a case study focused on digitized manuscripts from the Jagiellonian Digital Library.


\section{Case Study: Incunabula from the Jagiellonian Digital Library}

In order to validate the MEM framework, we conducted a preliminary study with usage of a dataset of digitized incunabula from the Jagiellonian Digital Library. This section present the dataset’s creation, implementation steps, and challenges encountered during the process.

\subsection{Dataset and Implementation}
\begin{figure}[!ht]
    \centering
    \includegraphics[width=0.4\textwidth]{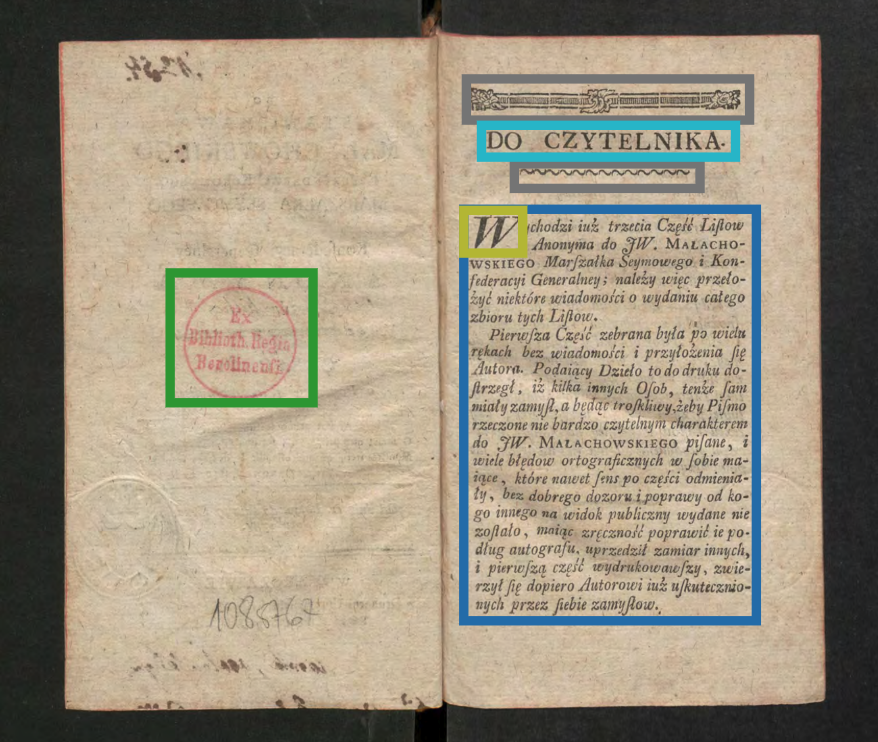}
    \caption{An example image from the created dataset. The image shows the labeled classes: \textit{stamp}, \textit{initial}, \textit{header}, \textit{ornament}, and \textit{paragraph}.}
    \label{fig:example_dataset_image}
\end{figure}

A manually annotated dataset consisting of 105 manuscript pages was created specifically for this work. Visual elements were marked across ten distinct categories, such as: Paragraph, Ornament, Stain, Stamp, Header, Initial, Seal, Signature, Image, Description. The dataset is publicly released under a CC-BY license here: https://www.kaggle.com/datasets/janignatowicz/105-jdl-labelled-incunabula-images.
An example image from the dataset with labeled classes is presented in Figure~\ref{fig:example_dataset_image}.

\begin{figure*}[t]
    \centering
    \includegraphics[width=\textwidth]{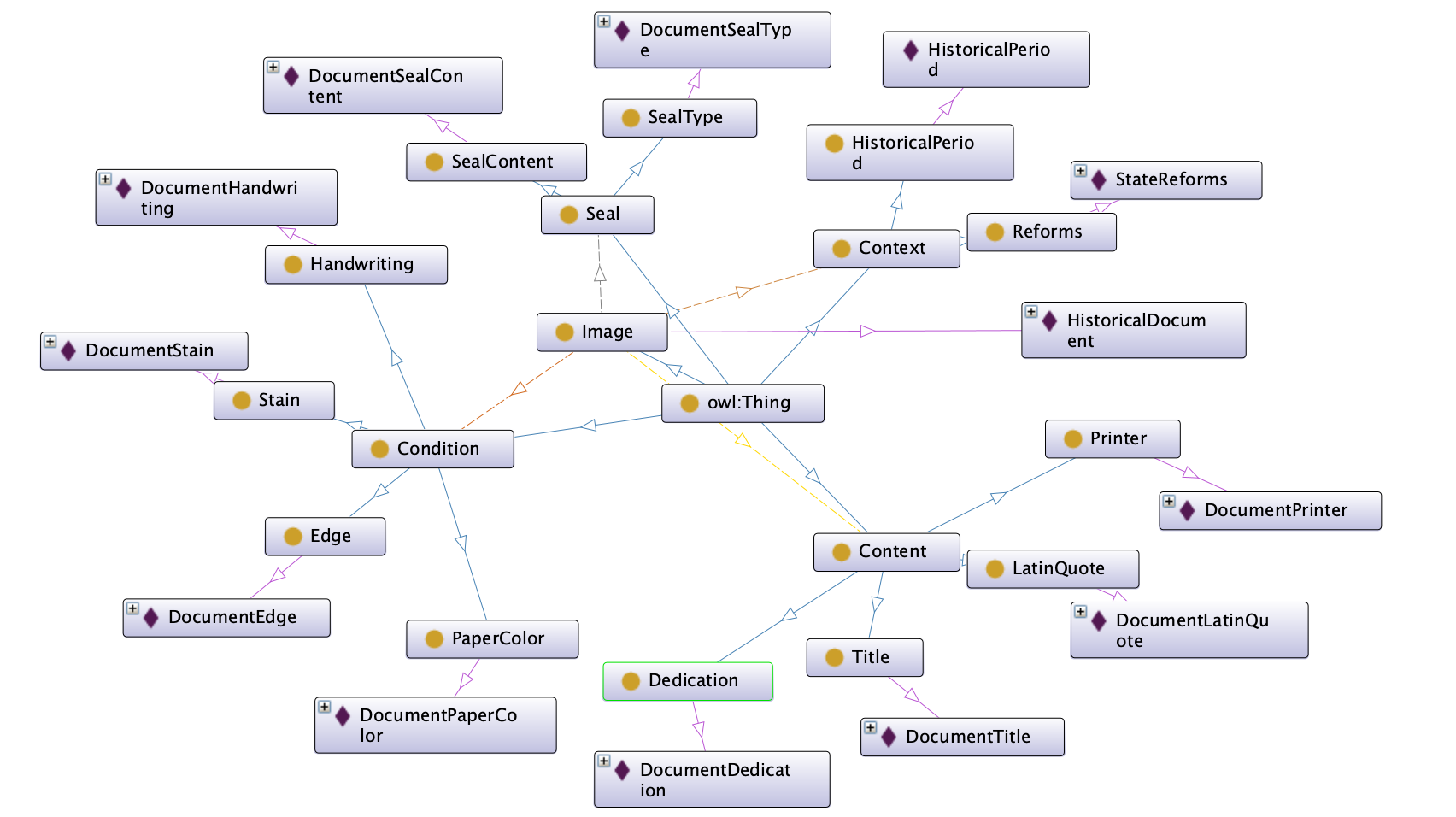}
    \caption{Example created ontology structure based on found new metadata and provided metadata by institution.}
    \label{fig:ontology_structure}
\end{figure*}

For object detection, we adapted two widely used models (YOLOv11 and Detectron2) by training them on a custom annotated dataset designed to recognize ten distinct objects found in manuscripts. Both models showed the ability to detect desired visual elements. However, challenges arose in distinguishing complex ornamentations from text regions. A draft RDF ontology was created to structure the extracted metadata, focusing on contextual links (e.g.\textit{mem:contains}, \textit{mem:inConditionState}). An example created ontology structure, that was visualized using Protégé desktop app, is presented in Figure~\ref{fig:ontology_structure}.

\subsection{Challenges}
Base challanges appear with annotation complexity, models' limitations and struturing the ontologies. Subtle variations in decorative elements (e.g. floral vs. geometric patterns) require iterative revisions for enhancing labeling guidelines. The models detected most high-level features, but occasionally misclassified destroyed or overlapping elements (e.g. seals as decorative borders).

\subsection{Future Work}
Those preliminary results validated the usefulness of MEM. Future work should focus on scaling up the framework, by integration multimodal data, and clarify the ontology with community input.


\section{Discussion}

The Metadata Enrichment Model (MEM) propose a change in how cultural heritage institutions could approach metadata retrieval. With integration of neural networks, semantic technologies, and iterative analysis, MEM presents a step forward in metadata enrichment processes. Particularly, by describing artifacts with nested or multimodal features. Below, we indicate MEM’s contributions, limitations, and implications for future research.

\subsection{AI with Semantic Technologies}
MEM’s innovation lies in its ability to dynamically link AI-driven insights with structured knowledge graphs. Unlike static metadata schemas, MEM’s RDF-based ontology may evolve with new retrieved information. For example, automatically defining a \textit{mem:SealWithText} class, when text is detected within a seal. This flexibility meets the special needs of cultural heritage collections. However, adjusting domain-specific terms with external databases like Wikidata remains as a challenge, what implies a need for expert-based collaborative ontology development.

\subsection{Implications for GLAM Institutions}
For galleries, libraries, archives, and museums (GLAM) MEM presents a scalable and modular extension to manual metadata retrieval. The framework’s design allows institutions to implement components incrementally. Institutions may start with object detection and then integrate visual analysis results with LLMs. The release of our annotated dataset indicate directions for further dataset enlargements, enabling institutions to retrain models. A significant obstacle limiting implementations in underfunded organizations may turn out to be MEM’s computational demands, like GPU requirements for retraining computer vision models.

\subsection{Ethical and Methodological Considerations}
The automation of metadata enrichment presents ethical dilemmas, particularly around cultural representation and algorithmic biases. When AI models are trained predominantly i.e. on Eurocentric datasets, they risk of misinterpreting or even erasing non-Western cultural expressions, including iconography, symbolism, and contextual narratives. Such behaviour may contribute to perpetuation historical inequities in digital heritage preservation, where marginalized cases are already underrepresented.

To overcome these obstacles, future metadata enrichment systems should adopt some safeguards. Firstly, they must focus on transparency by logging model decisions comprehensively, ensuring auditability and accountability in automated metadata generation. 

Secondly, technical solutions alone cannot adequately solve the problem of existing errors in such systems. Here, human verification is needed to claim whether automated approaches might unintentionally reinforce existing archival structures. Implementing human-in-the-loop validation mechanisms would help with ensurance that ontologies and classifications remain in domain expertise, not in purely algorithmic outputs. Development processes of such systems should involve contributions from historians, ethicists, and community stakeholders in order to establish more culturally inclusive definitions of accuracy.

Future research should focus on two priorities: diversifying training datasets to better represent marginalized cultural contexts (countering homogenization), and collaboratively addressing data biases through partnerships with source communities. 

\subsection{Limitations and Future Directions}
MEM shows prossible directions, but several limitations still need attention. The quality of results depends on quality of labeled data and their amount, which is still lacking in niche datasets. Processing time per image may hinder large-scale deployments as the scalability limitation.

Future research may focus on: multimodal expansion, by integrating 3D scans, audio descriptions, and geospatial data; as well as on community-driven ontologies by partnering with historians and librarians to clarify semantic schemas; and eventually on developing guidelines for AI-assisted cultural heritage retrieval. An interesting direction will be the  research for other fields that may use MEMin order to enrich metadata.

\section{Conclusion}

This paper presents the concept of Metadata Enrichment Model (MEM), a modular framework designed to enrich metadata. Presented in applications for digitized cultural heritage collections through the integration of neural networks and semantic technologies MEM allows for iterative analysis of nested features by its Multilayer Vision Mechanism (MVM) module. Moreover it dynamically structures findings, such as text in seals, in RDF-based knowledge graphs. The Jagiellonian Digital Library of incunabulum use case showed MEM’s potential, although some challenges in model generalization and ontology adjustment remain. 

MEM is not a replacement for human expertise but a tool to strenghten it. As cultural heritage digitization accelerates, frameworks like MEM will become more and more important to retrieve historical assets, enabling cross-collection research, and faciliating global access. We invite the community to continue upon this work, using the released dataset to explore new frontiers in AI-driven heritage retrieval cases.


\section*{Acknowledgment}
The research has been supported by a grant from the Priority Research Area (DigiWorld) under the Strategic Programme Excellence Initiative at Jagiellonian University. This publication was funded by a flagship project "CHExRISH: Cultural Heritage Exploration and Retrieval with Intelligent Systems at Jagiellonian University" under the Strategic Programme Excellence Initiative at Jagiellonian University.

\bibliographystyle{IEEEtran}
\bibliography{bibliography}

\begin{thebibliography}{10}
\providecommand{\url}[1]{#1}
\csname url@samestyle\endcsname
\providecommand{\newblock}{\relax}
\providecommand{\bibinfo}[2]{#2}
\providecommand{\BIBentrySTDinterwordspacing}{\spaceskip=0pt\relax}
\providecommand{\BIBentryALTinterwordstretchfactor}{4}
\providecommand{\BIBentryALTinterwordspacing}{\spaceskip=\fontdimen2\font plus
\BIBentryALTinterwordstretchfactor\fontdimen3\font minus
  \fontdimen4\font\relax}
\providecommand{\BIBforeignlanguage}[2]{{%
\expandafter\ifx\csname l@#1\endcsname\relax
\typeout{** WARNING: IEEEtran.bst: No hyphenation pattern has been}%
\typeout{** loaded for the language `#1'. Using the pattern for}%
\typeout{** the default language instead.}%
\else
\language=\csname l@#1\endcsname
\fi
#2}}
\providecommand{\BIBdecl}{\relax}
\BIBdecl

\bibitem{vanhooland_facet_2014}
\BIBentryALTinterwordspacing
S.~van Hooland and R.~Verborgh, \emph{Linked Data for Libraries, Archives and
  Museums}.\hskip 1em plus 0.5em minus 0.4em\relax Facet Publishing, Jun. 2014.
  [Online]. Available: \url{http://amzn.to/UUfjfN}
\BIBentrySTDinterwordspacing

\bibitem{10.1145/3607542.3617354}
\BIBentryALTinterwordspacing
V.~de~Boer, ``Knowledge graphs for cultural heritage and digital humanities,''
  in \emph{Proceedings of the 5th Workshop on AnalySis, Understanding and
  ProMotion of HeritAge Contents}, ser. SUMAC '23.\hskip 1em plus 0.5em minus
  0.4em\relax New York, NY, USA: Association for Computing Machinery, 2023,
  p.~3. [Online]. Available: \url{https://doi.org/10.1145/3607542.3617354}
\BIBentrySTDinterwordspacing

\bibitem{Weibel1998}
S.~Weibel, J.~Kunze, C.~Lagoze, and M.~Wolf, ``{Dublin Core Metadata for
  Resource Discovery},'' Internet Engineering Task Force (IETF), Tech. Rep. RFC
  2413, 1998.

\bibitem{guenther2003marc}
\BIBentryALTinterwordspacing
R.~Guenther, ``Marc 21 format for bibliographic data,'' \emph{Library of
  Congress}, 2003. [Online]. Available:
  \url{https://www.loc.gov/marc/bibliographic/}
\BIBentrySTDinterwordspacing

\bibitem{Doerr2003}
M.~Doerr, ``{The CIDOC conceptual reference model: An ontological approach to
  semantic interoperability of metadata},'' \emph{AI Magazine}, vol.~24, no.~3,
  pp. 75--92, 2003.

\bibitem{berners2006linked}
T.~Berners-Lee, J.~Hendler, and O.~Lassila, ``The semantic web,''
  \emph{Scientific American}, vol. 284, no.~5, pp. 34--43, 2001.

\bibitem{10.1007/s00371-024-03343-0}
\BIBentryALTinterwordspacing
H.~Senior, G.~Slabaugh, S.~Yuan, and L.~Rossi, ``Graph neural networks in
  vision-language image understanding: a survey: Graph neural networks in
  vision-language image understanding: a survey,'' \emph{Vis. Comput.},
  vol.~41, no.~1, p. 491–516, Mar. 2024. [Online]. Available:
  \url{https://doi.org/10.1007/s00371-024-03343-0}
\BIBentrySTDinterwordspacing

\bibitem{Khanam2024}
\BIBentryALTinterwordspacing
R.~Khanam and M.~Hussain, ``Yolov11: An overview of the key architectural
  enhancements,'' \emph{arXiv preprint arXiv:2410.17725}, 2024. [Online].
  Available: \url{https://arxiv.org/abs/2410.17725}
\BIBentrySTDinterwordspacing

\bibitem{wu2019detectron2}
Y.~Wu, A.~Kirillov, F.~Massa, W.-Y. Lo, and R.~Girshick, ``Detectron2,''
  \url{https://github.com/facebookresearch/detectron2}, 2019.

\bibitem{wang2020deep}
\BIBentryALTinterwordspacing
J.~Wang, K.~Sun, T.~Cheng, B.~Jiang, C.~Deng, Y.~Zhao, D.~Liu, Y.~Mu, M.~Tan,
  X.~Wang, W.~Liu, and B.~Xiao, ``Deep high-resolution representation learning
  for visual recognition,'' \emph{IEEE Transactions on Pattern Analysis and
  Machine Intelligence}, 2020. [Online]. Available:
  \url{https://arxiv.org/abs/1908.07919}
\BIBentrySTDinterwordspacing

\bibitem{bommasani2021opportunities}
R.~Bommasani, D.~A. Hudson, E.~Adeli, R.~Altman, S.~Arora, S.~von Arx, M.~S.
  Bernstein, J.~Bohg, A.~Bosselut, E.~Brunskill \emph{et~al.}, ``On the
  opportunities and risks of foundation models,'' \emph{arXiv preprint
  arXiv:2108.07258}, 2021.

\bibitem{Garcia-Garcia2017}
\BIBentryALTinterwordspacing
A.~Garcia{-}Garcia, S.~Orts{-}Escolano, S.~Oprea, V.~Villena{-}Martinez, and
  J.~Garcia{-}Rodriguez, ``A review on deep learning techniques applied to
  semantic segmentation,'' \emph{CoRR}, vol. abs/1704.06857, 2017. [Online].
  Available: \url{http://arxiv.org/abs/1704.06857}
\BIBentrySTDinterwordspacing

\bibitem{shi2023gold}
\BIBentryALTinterwordspacing
C.~Liu, S.~Wang, L.~Qing, K.~Kuang, Y.~Kang, C.~Sun, and F.~Wu, ``Gold panning
  in vocabulary: An adaptive method for vocabulary expansion of domain-specific
  llms,'' 2024. [Online]. Available: \url{https://arxiv.org/abs/2410.01188}
\BIBentrySTDinterwordspacing

\bibitem{liu2024knowledge}
\BIBentryALTinterwordspacing
H.~Liu, S.~Wang, Y.~Zhu, Y.~Dong, and J.~Li, ``Knowledge graph-enhanced large
  language models via path selection,'' 2024. [Online]. Available:
  \url{https://arxiv.org/abs/2406.13862}
\BIBentrySTDinterwordspacing

\bibitem{isaac2009europeana}
\BIBentryALTinterwordspacing
A.~Isaac and B.~Haslhofer, ``Europeana data model primer,'' \emph{Europeana
  Foundation}, 2009. [Online]. Available:
  \url{https://pro.europeana.eu/files/Europeana_Professional/Share_your_data/Technical_requirements/EDM_Documentation/EDM_Primer_130714.pdf}
\BIBentrySTDinterwordspacing

\bibitem{chen2017rethinking}
\BIBentryALTinterwordspacing
L.-C. Chen, G.~Papandreou, I.~Kokkinos, K.~Murphy, and A.~L. Yuille,
  ``Rethinking atrous convolution for semantic image segmentation,''
  \emph{arXiv preprint arXiv:1706.05587}, 2017. [Online]. Available:
  \url{https://arxiv.org/abs/1706.05587}
\BIBentrySTDinterwordspacing

\bibitem{openai2023gpt4}
\BIBentryALTinterwordspacing
OpenAI, ``Gpt-4 technical report,'' \emph{arXiv preprint arXiv:2303.08774},
  2023. [Online]. Available: \url{https://arxiv.org/abs/2303.08774}
\BIBentrySTDinterwordspacing

\bibitem{hong2024semantic}
\BIBentryALTinterwordspacing
Z.~Chen, Y.~Zhang, Y.~Fang, Y.~Geng, L.~Guo, X.~Chen, Q.~Li, W.~Zhang, J.~Chen,
  Y.~Zhu, J.~Li, X.~Liu, J.~Z. Pan, N.~Zhang, and H.~Chen, ``Knowledge graphs
  meet multi-modal learning: A comprehensive survey,'' 2024. [Online].
  Available: \url{https://arxiv.org/abs/2402.05391}
\BIBentrySTDinterwordspacing

\bibitem{smith2023ocr}
P.~Maldonado-Quispe and H.~Pedrini, ``An effective approach to text detection
  and recognition in degraded historical documents,'' in \emph{Progress in
  Pattern Recognition, Image Analysis, Computer Vision, and Applications},
  R.~Hern{\'a}ndez-Garc{\'i}a, R.~J. Barrientos, and S.~A. Velastin, Eds.\hskip
  1em plus 0.5em minus 0.4em\relax Cham: Springer Nature Switzerland, 2025, pp.
  256--269.

\bibitem{vrandecic2014wikidata}
D.~Vrandečić and M.~Krötzsch, ``Wikidata: A free collaborative
  knowledgebase,'' \emph{Communications of the ACM}, vol.~57, no.~10, pp.
  78--85, 2014.

\bibitem{wilkinson2016fair}
M.~D. Wilkinson \emph{et~al.}, ``The fair guiding principles for scientific
  data management,'' \emph{Scientific Data}, vol.~3, p. 160018, 2016.

\bibitem{jdl}
{Jagiellonian University}, ``Jagiellonian digital library,''
  \url{https://jbc.bj.uj.edu.pl/dlibra}, 2023, [Online; accessed 20-June-2023].

\bibitem{tan2019efficientnet}
\BIBentryALTinterwordspacing
M.~Tan and Q.~V. Le, ``Efficientnet: Rethinking model scaling for convolutional
  neural networks,'' 2020. [Online]. Available:
  \url{https://arxiv.org/abs/1905.11946}
\BIBentrySTDinterwordspacing

\bibitem{vanhooland_llc_2015}
\BIBentryALTinterwordspacing
S.~van Hooland, M.~De~Wilde, R.~Verborgh, T.~Steiner, and R.~Van~de Walle,
  ``Exploring entity recognition and disambiguation for cultural heritage
  collections,'' \emph{Digital Scholarship in the Humanities}, vol.~30, no.~2,
  pp. 262--279, Jun. 2015. [Online]. Available:
  \url{http://freeyourmetadata.org/publications/named-entity-recognition.pdf}
\BIBentrySTDinterwordspacing

\bibitem{he2016deep}
K.~He \emph{et~al.}, ``Deep residual learning for image recognition,''
  \emph{CVPR}, pp. 770--778, 2016.

\bibitem{BernersLee2006}
\BIBentryALTinterwordspacing
{Tim Berners-Lee}, ``{Linked Data - Design Issues},'' 2006. [Online].
  Available: \url{https://www.w3.org/DesignIssues/LinkedData.html}
\BIBentrySTDinterwordspacing

\bibitem{Purday2009}
J.~Purday, ``{Think culture: Europeana.eu from concept to construction},''
  \emph{The Electronic Library}, vol.~27, no.~6, pp. 919--937, 2009.

\bibitem{knowledge-graph-visualization}
\BIBentryALTinterwordspacing
C.~S. Khoo, E.~A. Tan, S.-G. Ng, C.-F. Chan, M.~Stanley-Baker, and W.-N. Cheng,
  ``Knowledge graph visualization interface for digital heritage collections:
  Design issues and recommendations,'' \emph{Information Technology and
  Libraries}, vol.~43, no.~1, Mar. 2024. [Online]. Available:
  \url{https://ital.corejournals.org/index.php/ital/article/view/16719}
\BIBentrySTDinterwordspacing

\bibitem{ollama}
\BIBentryALTinterwordspacing
{Ollama Team}, ``{Ollama: Run Large Language Models Locally},'' 2025. [Online].
  Available: \url{https://github.com/ollama/ollama}
\BIBentrySTDinterwordspacing

\bibitem{deepseekv3}
\BIBentryALTinterwordspacing
{DeepSeek AI Team}, ``{DeepSeek v3: Advanced AI Language Model},'' 2025.
  [Online]. Available: \url{https://github.com/deepseek-ai/DeepSeek-V3}
\BIBentrySTDinterwordspacing

\bibitem{10.1007/978-3-031-77844-5_9}
G.~Pons, B.~Bilalli, and A.~Queralt, ``Knowledge graphs for enhancing large
  language models in entity disambiguation,'' in \emph{The Semantic Web --
  ISWC 2024}, G.~Demartini, K.~Hose, M.~Acosta, M.~Palmonari, G.~Cheng,
  H.~Skaf-Molli, N.~Ferranti, D.~Hern{\'a}ndez, and A.~Hogan, Eds.\hskip 1em
  plus 0.5em minus 0.4em\relax Cham: Springer Nature Switzerland, 2025, pp.
  162--179.

\bibitem{chepurova-etal-2023-better}
\BIBentryALTinterwordspacing
A.~Chepurova, A.~Bulatov, Y.~Kuratov, and M.~Burtsev, ``Better together:
  Enhancing generative knowledge graph completion with language models and
  neighborhood information,'' in \emph{Findings of the Association for
  Computational Linguistics: EMNLP 2023}, H.~Bouamor, J.~Pino, and K.~Bali,
  Eds.\hskip 1em plus 0.5em minus 0.4em\relax Singapore: Association for
  Computational Linguistics, Dec. 2023, pp. 5306--5316. [Online]. Available:
  \url{https://aclanthology.org/2023.findings-emnlp.352/}
\BIBentrySTDinterwordspacing

\bibitem{Lippolis2023EnhancingEA}
\BIBentryALTinterwordspacing
A.~S. Lippolis, A.~Klironomos, D.~F. Milon-Flores, H.~Zheng, A.~Jouglar,
  E.~Norouzi, and A.~Hogan, ``Enhancing entity alignment between wikidata and
  artgraph using llms,'' in \emph{International Joint Workshop on Semantic Web
  and Ontology Design for Cultural Heritage}, 2023. [Online]. Available:
  \url{https://api.semanticscholar.org/CorpusID:263913092}
\BIBentrySTDinterwordspacing

\bibitem{d3159688d57f4176808086ec1de5377a}
B.~Zhang, I.~Reklos, N.~Jain, A.~Pe{\~n}uela, and E.~Simperl,
  ``\BIBforeignlanguage{English}{Using large language models for knowledge
  engineering (llmke): A case study on wikidata},''
  \emph{\BIBforeignlanguage{English}{CEUR Workshop Proceedings}}, vol. 3577,
  Jan. 2023.

\bibitem{lairgi:hal-04607294}
\BIBentryALTinterwordspacing
Y.~Lairgi, L.~Moncla, R.~Cazabet, K.~Benabdeslem, and P.~Cl{\'e}au,
  ``{Knowledge Graph Construction Using Large Language Models},'' in
  \emph{{Journ{\'e}e nationale sur la fouille de textes}}, Lyon, France, Jun.
  2024. [Online]. Available: \url{https://hal.science/hal-04607294}
\BIBentrySTDinterwordspacing

\bibitem{SHIMIZU2025100862}
\BIBentryALTinterwordspacing
C.~Shimizu and P.~Hitzler, ``Accelerating knowledge graph and ontology
  engineering with large language models,'' \emph{Journal of Web Semantics}, p.
  100862, 2025. [Online]. Available:
  \url{https://www.sciencedirect.com/science/article/pii/S1570826825000022}
\BIBentrySTDinterwordspacing

\bibitem{10.1145/3581783.3611753}
\BIBentryALTinterwordspacing
C.~Fang, J.~Li, L.~Li, C.~Ma, and D.~Hu, ``Separate and locate: Rethink the
  text in text-based visual question answering,'' in \emph{Proceedings of the
  31st ACM International Conference on Multimedia}, ser. MM '23.\hskip 1em plus
  0.5em minus 0.4em\relax New York, NY, USA: Association for Computing
  Machinery, 2023, p. 4378–4388. [Online]. Available:
  \url{https://doi.org/10.1145/3581783.3611753}
\BIBentrySTDinterwordspacing

\bibitem{10.1609/aaai.v37i9.26357}
\BIBentryALTinterwordspacing
Y.~Zhu, Z.~Liu, Y.~Liang, X.~Li, H.~Liu, C.~Bao, and L.~Xu, ``Locate then
  generate: bridging vision and language with bounding box for scene-text
  vqa,'' in \emph{Proceedings of the Thirty-Seventh AAAI Conference on
  Artificial Intelligence and Thirty-Fifth Conference on Innovative
  Applications of Artificial Intelligence and Thirteenth Symposium on
  Educational Advances in Artificial Intelligence}, ser.
  AAAI'23/IAAI'23/EAAI'23.\hskip 1em plus 0.5em minus 0.4em\relax AAAI Press,
  2023. [Online]. Available: \url{https://doi.org/10.1609/aaai.v37i9.26357}
\BIBentrySTDinterwordspacing

\bibitem{britishlibrary2011bnb}
\BIBentryALTinterwordspacing
B.~Library, ``Publishing the british national bibliography as linked open
  data,'' \emph{British Library}, 2011. [Online]. Available:
  \url{https://www.bl.uk/bibliographic/pdfs/british_national_bibliography_linked_open_data.pdf}
\BIBentrySTDinterwordspacing

\bibitem{liu-etal-2024-knowledge-graph}
\BIBentryALTinterwordspacing
H.~Liu, S.~Wang, Y.~Zhu, Y.~Dong, and J.~Li, ``Knowledge graph-enhanced large
  language models via path selection,'' in \emph{Findings of the Association
  for Computational Linguistics: ACL 2024}, L.-W. Ku, A.~Martins, and
  V.~Srikumar, Eds.\hskip 1em plus 0.5em minus 0.4em\relax Bangkok, Thailand:
  Association for Computational Linguistics, Aug. 2024, pp. 6311--6321.
  [Online]. Available: \url{https://aclanthology.org/2024.findings-acl.376/}
\BIBentrySTDinterwordspacing

\bibitem{li2024handwritingrecognitionhistoricaldocuments}
\BIBentryALTinterwordspacing
L.~Li, ``Handwriting recognition in historical documents with multimodal llm,''
  2024. [Online]. Available: \url{https://arxiv.org/abs/2410.24034}
\BIBentrySTDinterwordspacing

\bibitem{gao2023enabling}
T.~Gao, H.~Yen, J.~Yu, and D.~Chen, ``Enabling large language models to
  generate text with citations,'' in \emph{Proceedings of the 2023 Conference
  on Empirical Methods in Natural Language Processing}, 2023, pp. 6465--6488.

\bibitem{zhang2024longcite}
J.~Zhang \emph{et~al.}, ``Enabling llms to generate fine-grained citations in
  long-context qa,'' \emph{arXiv preprint arXiv:2409.02897}, 2024.

\bibitem{kazemi2025refining}
M.~Kazemi~Rad \emph{et~al.}, ``Refining input guardrails: Enhancing
  llm-as-a-judge efficiency through chain-of-thought fine-tuning and
  alignment,'' \emph{arXiv preprint arXiv:2501.13080}, 2025.

\bibitem{li2024rulerimprovingllmcontrollability}
M.~Li \emph{et~al.}, ``Ruler: Improving llm controllability by rule-based data
  recycling,'' \emph{arXiv preprint arXiv:2406.15938}, 2024.

\end{thebibliography}

\vspace{12pt}

\end{document}